\documentclass[10pt, a4paper]{article}
\usepackage{lrec-coling2024} 

\usepackage{helvet}

\usepackage{natbib}
\usepackage{multibib}
\makeatletter
\def\@mb@citenamelist{cite,citep,citet,citealp,citealt,citepalias,citetalias}
\makeatother
\newcites{languageresource}{~}

\usepackage{graphicx}
\usepackage{tabularx}
\usepackage{soul}
\usepackage{titlesec}
\titleformat{\section}{\normalfont\large\bfseries\center}{\thesection.}{1em}{}
\titleformat{\subsection}{\normalfont\SmallTitleFont\bfseries\raggedright}{\thesubsection.}{1em}{}
\titleformat{\subsubsection}{\normalfont\normalsize\bfseries\raggedright}{\thesubsubsection.}{1em}{}
\renewcommand\thesection{\arabic{section}}
\renewcommand\thesubsection{\thesection.\arabic{subsection}}
\renewcommand\thesubsubsection{\thesubsection.\arabic{subsubsection}}

\usepackage{xcolor}
\usepackage{hyperref}
 \definecolor{darkblue}{rgb}{0, 0, 0.5}
  \hypersetup{colorlinks=true, citecolor=darkblue, linkcolor=darkblue, urlcolor=darkblue}

\usepackage{xstring}

\usepackage{color}
\usepackage{amsmath}
\usepackage{booktabs}
\usepackage{makecell}
\usepackage{multirow}

\title{Dynamic Spatial-Temporal Aggregation for Skeleton-Aware Sign Language Recognition}

\name{Lianyu Hu, Liqing Gao, Zekang Liu, Wei Feng$^*$\thanks{Corresponding author}} 

\address{College of Intelligence and Computing, Tianjin University, Tianjin 300350, China \\
         \{hly2021, lqgao,
         lzk100953, wfeng\}@tju.edu.cn\\}

\abstract{
Skeleton-aware sign language recognition (SLR) has gained popularity due to its ability to remain unaffected by background information and its lower computational requirements. Current methods utilize spatial graph modules and temporal modules to capture spatial and temporal features, respectively. However, their spatial graph modules are typically built on fixed graph structures such as graph convolutional networks or a single learnable graph, which only partially explore joint relationships. Additionally, a simple temporal convolution kernel is used to capture temporal information, which may not fully capture the complex movement patterns of different signers. To overcome these limitations, we propose a new spatial architecture consisting of two concurrent branches, which build input-sensitive joint relationships and incorporates specific domain knowledge for recognition, respectively. These two branches are followed by an aggregation process to distinguishe important joint connections. We then propose a new temporal module to model multi-scale temporal information to capture complex human dynamics. Our method achieves state-of-the-art accuracy compared to previous skeleton-aware methods on four large-scale SLR benchmarks. Moreover, our method demonstrates superior accuracy compared to RGB-based methods in most cases while requiring much fewer computational resources, bringing better accuracy-computation trade-off. Code is available at \url{https://github.com/hulianyuyy/DSTA-SLR} 
 \\ \newline  }

\begin{document}

\maketitleabstract

\section{Introduction}
Sign language is the primary means of communication for deaf individuals, conveyed through dynamic hand gestures, body posture, and facial expressions~\cite{dreuw2007speech,ong2005automatic}. However, understanding sign language can be challenging and time-consuming for the hearing people, requiring significant time and effort. Fortunately, recent advancements in machine learning and computer vision have made significant progress in the automatic interpretation of sign language. These technologies greatly facilitate communication for deaf individuals, enabling them to interact more smoothly with others in their daily lives. 

Sign language recognition (SLR) can be broadly categorized into two types: vision-based SLR and skeleton-aware SLR. Vision-based SLR directly predicts signs from RGB image streams, which can be computationally expensive. Skeleton-aware SLR involves predicting a single sign label from a sequence of 2D or 3D skeleton representations. Compared to vision-based SLR, skeleton-aware SLR requires much fewer computations from 2D~\cite{kay2017kinetics,yan2018spatial} or 3D~\cite{liu2019ntu,shahroudy2016ntu} skeleton inputs and is less affected by environmental factors like camera motion, lighting conditions, and changes in viewpoint~\cite{liu2020disentangling}, which is thus more robust in extracting human action representations. 

Current methods for skeleton-aware SLR~\cite{tunga2021pose,jiang2021skeleton,bohavcek2022sign} often use a combination of spatial graph modules and temporal modules to capture spatial and temporal features, respectively. However, there are two main limitations to these methods. Firstly, current methods rely on graph convolutional networks (GCNs) with a fixed graph structure or a single learnable graph, which can hinder the modeling of dynamic connections between human body joints that are specific to different input samples. Secondly, current methods typically use a simple temporal convolution to capture temporal information, which may not be able to effectively model the complex temporal dependencies present in sign language. 

In this paper, we address the aforementioned limitations from two aspects. First, we propose to enhance the flexibility of spatial modules to comprehensively capture the correlations between different body joints. Specifically, we introduce a graph correlation module that dynamically builds spatial graphs for each input sample in a channel-wise manner, eliminating the need for a predefined graph based on the human body structure. Furthermore, we introduce virtual super nodes in addition to the physical human body joints to incorporate specific domain knowledge into the model. An adaptive graph is then constructed to distinguish the inherently important connections between human joints. Secondly, to capture the complex temporal movements of various input samples, we propose a parallel temporal convolutional module to aggregate multi-scale temporal information. This module consists of a series of temporal convolutions with varying receptive fields, which enables capturing human dynamics across a wide range of temporal durations. 

With the advantages of high spatial and temporal capacity, our proposed method achieves new state-of-the-art accuracy on four commonly used SLR benchmarks. Especially, compared to vision-based methods which typically demand much higher computations and memory usage, our model achieves superior results in most cases with a better accuracy-computation trade-off, thereby bringing us one step closer towards real-life applications. Several ablation experiments confirm the effectiveness of our proposed modules.

\section{Related Work}
\subsection{Sign Language Recognition}
Sign language recognition (SLR) aims to classify an input video into a single sign label, which can be roughly divided into vision-based SLR and skeleton-aware SLR. Vision-based SLR usually receives RGB videos or RGB+D videos as inputs. Hand pose is used as guidance to pool spatio-temporal feature maps from different layers of a 3D CNN in ~\cite{hosain2021hand} to capture multi-cue features. SignBERT~\cite{hu2021signbert} performs self-supervised pre-training on available sign data and incorporates hand prior in a model-aware method to model hierarchical context. SAM-SLR-v2~\cite{jiang2021sign} proposes to fuse the representations of multiple modalities to enhance the representations. Despite high accuracy, these vision-based methods usually require a lot of computations. Skeleton-aware SLR methods receive skeleton sequences as inputs to perform recognition, which are more invariant to background information and more lightweight in 2D or 3D format compared to RGB streams, thus requiring fewer computational resources. Previous skeleton-aware SLR methods mostly rely on graph networks to process non-Euclidean skeleton data by treating human body joints as graph nodes. SAM-SLR~\cite{jiang2021skeleton} proposes a sign language graph convolutional network and a separable spatial-temporal convolutional network to exploit skeleton features. SLGTformer~\cite{song2022slgtformer} leverages learnable graph relative positional encodings and temporal twin self-attention to guide skeleton feature extraction. SPOTER~\cite{bohavcek2022sign} introduces new normalization and augmentation techniques as well as a transformer-based framework to handle skeleton data. We argue that the graph structure adopted in these methods may not well fit complex human dynamics and propose to dynamically build graph structures. 

\subsection{Skeleton-based Action Recognition}
Skeleton-based action recognition aims to recognize human actions from a series of input skeleton sequences. Earlier methods always adopt convolutional neural networks (CNNs) or recurrent neural networks (RNNs) to depict action representations by modeling human dynamics as a pseudo-image~\cite{ke2017new,li2017skeleton,li2018co} or a series of coordinates along time~\cite{du2015hierarchical,liu2016spatio,li2018independently,zhang2017view,si2019attention}. Nevertheless, they overlook the internal relationships between joints which are better captured by graph networks due to their natural advantage over handling non-euclidean data. ST-GCN~\cite{yan2018spatial} has first proposed to model the human structure as a spatial graph by representing joints as nodes and bones as edges with a spatial-temporal graph convolutional network (GCN). However, the spatial receptive field in GCN is only limited to 1-hop neighbors and the edge weights are fixed in such conditions. Some late methods propose to leverage the higher polynomial order of the skeleton adjacency matrix to enlarge the spatial receptive field, where distant joints can exchange messages directly~\cite{gao2019optimized,li2019actional}. Some other methods propose to attach an attention-based adaptively computed graph along with the predefined graph to better exploit relationships between all joints and alleviate the predefined edge weights~\cite{shi2020skeleton,zhang2020semantics,xu2022topology,ke2022towards,su2021self,yang2021skeleton,chen2021channel}. 

\begin{figure}[t]
  \centering
  \includegraphics[width=0.6\columnwidth]{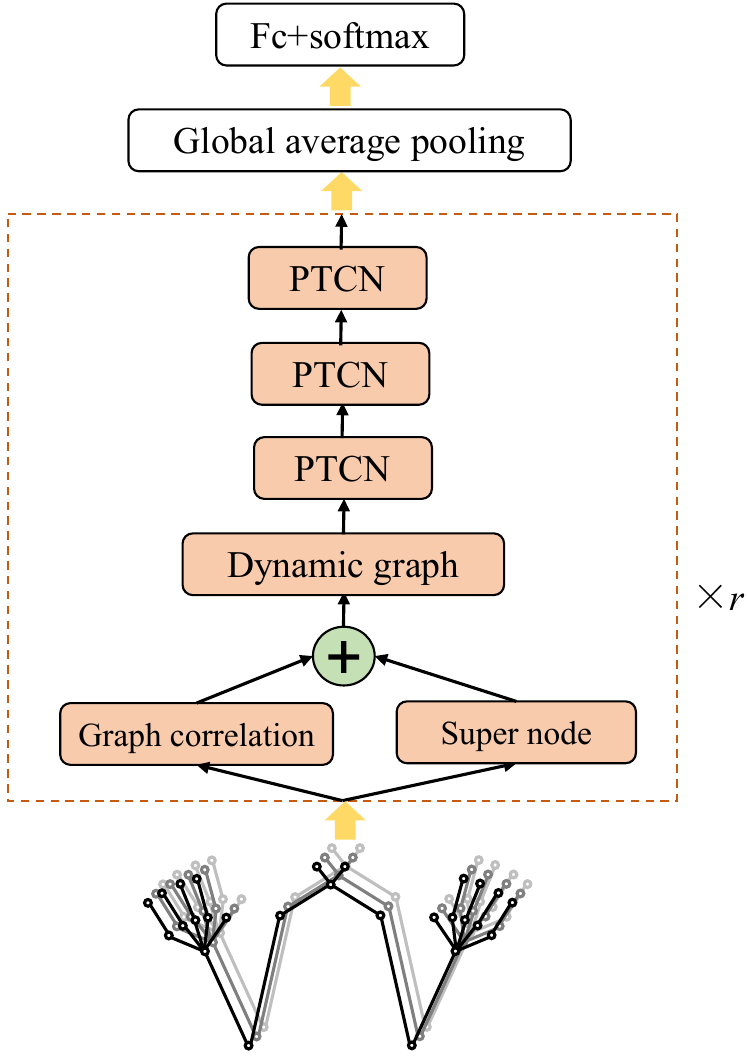} 
  \caption{Overview for our proposed model. } 
  \label{fig2}
  \end{figure}

\section{Methods}
Fig~\ref{fig2} illustrates the overview of our proposed method, which consists of $r$ blocks followed by a global average pooling layer, a fully connected (fc) layer, and a softmax function for recognition. Each block is first composed of a concurrent graph correlation module and a super node transform module to dynamically capture spatial joint relationships, followed by a dynamic graph aggregation module to distinguish inherently important joints. Three consecutive parallel temporal convolutional modules are followed to perform temporal modeling to capture complex human dynamics. In this section, we will first describe the process of building skeleton graphs and then provide a detailed explanation of our proposed method.

\subsection{Graph Construction}

Wearable motion capture devices such as Kinect V2~\cite{pagliari2015calibration} have been widely used in popular skeleton benchmarks such as NTU RGB+D 60~\cite{shahroudy2016ntu}, NTU RGB+D 120~\cite{liu2019ntu} and NW-UCLA~\cite{wang2014cross}. However, these devices cannot provide precise annotations for the hands, which are critical for SLR. To address this limitation, we use a pretrained whole-body pose estimation network (HRNet~\cite{sun2019deep} from MMPose~\cite{mmpose2020}) to offer annotations for people in the videos, resulting in 133 keypoints following the COCO format. To reduce the noise contained in the large number of nodes, we reduce the number of nodes to 27, which only contain 10 nodes for each hand and 7 nodes for the upper body, which is based on the observation that sign language is mainly conveyed by joints located in the upper human body. 

Given the input skeleton sequences, adjacent keypoints are connected in the spatial dimension according to the natural connections of the human body to form spatial graphs in GCNs. Formally, a graph is represented as \textit{G}=(\textit{V}, \textit{E}), where \textit{V} = \textit\{$v_{1}$,...,$v_{N}$\} is a set of \textit{N} graph nodes representing joints and \textit{E} is a series of graph edges representing connectivity (bones) between joints. An adjacent matrix \textit{A} with size \textit{N}$\times$\textit{N} is adopted to depict the connectivity where $A_{i,j}$ represents the connection strength between node \textit{i} and \textit{j}. Especially, $A$ is constructed as :
\begin{equation}
  \label{e1}
  \mathbf{A}_{i, j}= \begin{cases}1 & \text { if } d\left(v_i, v_j\right)=1 \\ 0 & \text { else }\end{cases}
\end{equation}
where d($v_i$, $v_j$) calculates the minimum distance between skeleton node $v_i$ and $v_j$. The action sequences are represented as a feature tensor $x\in\mathcal{R}^{C\times T\times N}$ where each node $v_{n}\in$ \textit{N} has a \textit{C} dimensional feature vector over total \textit{T} frames.

\begin{figure}[t]
  \centering
  \includegraphics[width=0.7\columnwidth]{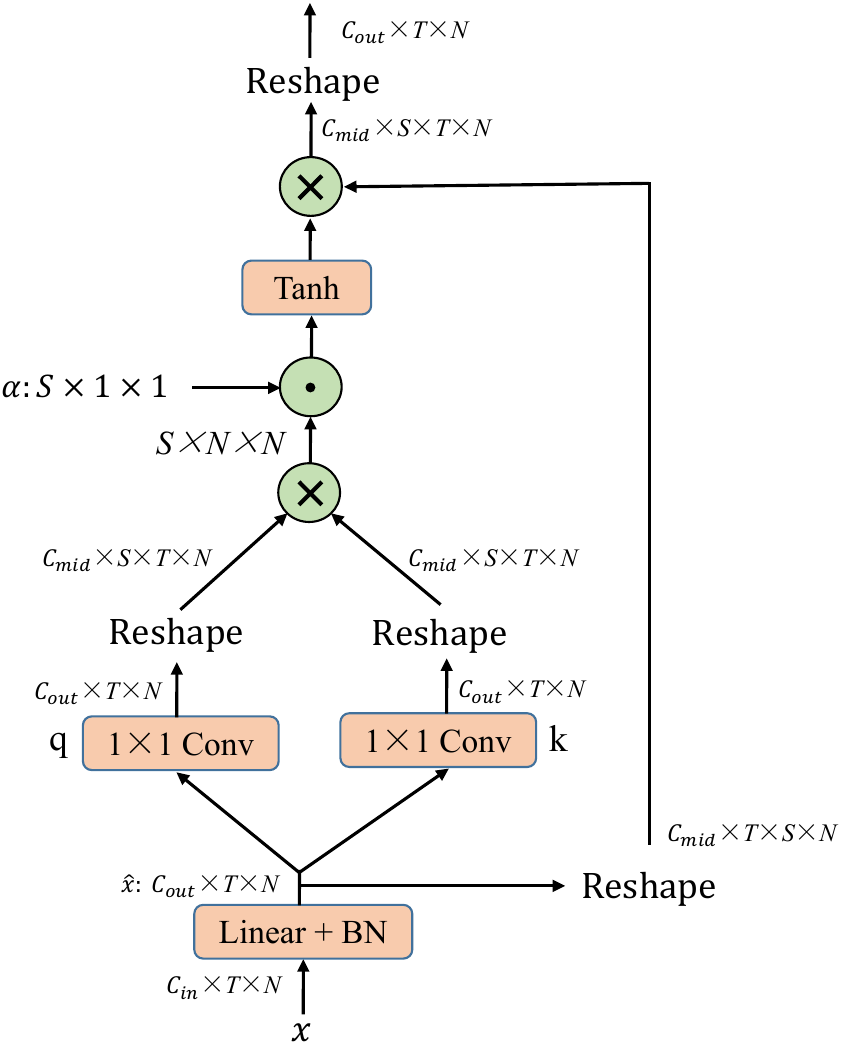} 
  \caption{Overview for the proposed graph correlation module.} 
  \label{fig3}
  \end{figure}

\subsection{Graph Correlation Module}
Recently, graph convolutional networks (GCNs) are broadly introduced to model human body dynamics in action recognition and SLR. Specifically, given the input tensor $x\in\mathcal{R}^{C\times T\times N}$, it performs layer-wise graph convolution with 1-hop normalized graph structure \textit{A} at each time step as:
\begin{equation}
\label{e2}
Y=\sigma(WXA)
\end{equation}    
where $Y$ is the output and \textit{W} is a trainable weight matrix. $\sigma (\cdot)$ acts as an activation function for output. In the graph convolutional network, the graph $A$ is predefined according to human body structure in eq.~\ref{e1} where each node can only aggregate information from 1-hop neighbors with fixed weights. However, distant nodes may still have significant correlations in sign language recognition, and these connections are not well captured by the fixed $A$. Moreover, the relationships between joints should be dynamically computed as different joints may play distinct roles in different sign language actions. We propose to dynamically compute joint relationships for each input sample to better capture important connections.

The overview of the graph correlation module is shown in fig.~\ref{fig3}. In each layer, the input features $x\in\mathcal{R}^{C_{in}\times T\times N}$ pass through a linear layer and a BatchNorm layer to obtain $\widehat{x}$, transforming the channels into $C_{out}$. Afterwards, the features from the query and key go through two 1$\times$1 convolutions to act as query and key, respectively. Then the features are reshaped into $\mathcal{R}^{C_{mid}\times S\times T\times N}$ to compute the adjacent matrices. In this process, $S$ independent adjacent matrices are generated to encode various joint relationships to expand the network capacity, where $C_{mid}$=$C_{out}/S$. Finally, the features from both branches are multiplied to produce the expected adjacent matrices $\widehat{A} \in \mathcal{R}^{S\times N\times N}$. To better represent the significance of each subset, we introduce a learnable parameter $\alpha$ that is randomly initialized and multiplied with the computed adjacent matrices. The resulting ajacent matrices $\widehat{A}$ are passed through a tanh activation function to transform their values into the range of [-1, 1]. Negative values in $\widehat{A}$ are expected to suppress corresponding messages, while positive values are expected to aggregate useful information from corresponding nodes. The reshaped $\widehat{x}$ is finallt multiplied with the adjacent matrices $\widehat{A}$ to aggregate information from other body joints. It's worth noting that in this process, the adjacent matrices are dynamically and independently computed for each input sample, getting rid of the predefined pattern of previous methods. Thus, it could dynamically build relationships with distant nodes, avoiding fixed weights within only 1-hop neighbors. 

\subsection{Super Node Transform Module}
\begin{figure}[t]
  \centering
  \includegraphics[width=0.6\columnwidth]{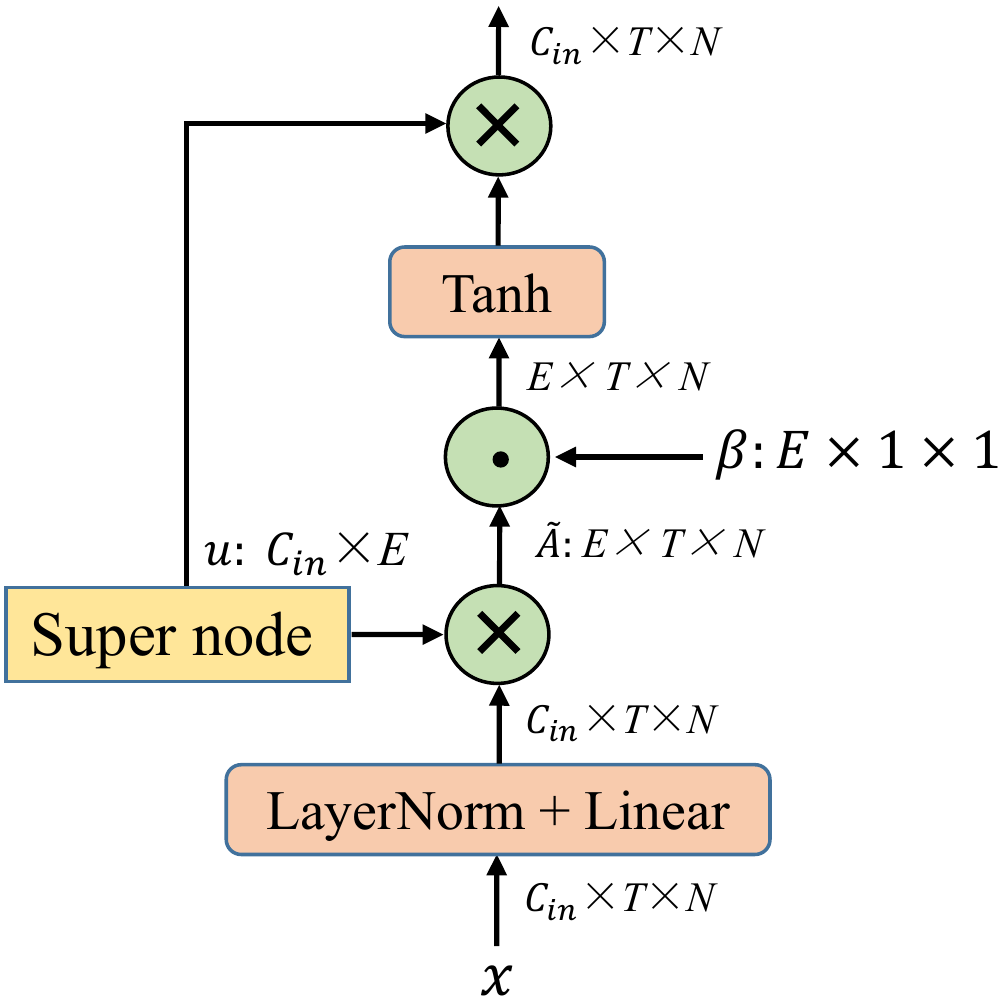} 
  \caption{Overview for the proposed super node transform module.} 
  \label{fig4}
\end{figure}

Currently, only human body joints are assigned as graph nodes, with specific physical meanings in practice. However, we argue that the network could also contain task-specific knowledge inherently to aid recognition. For instance, in SLR, we expect the network to contain the rules to perform signs and execute body movements to help recognition. To achieve this, we propose adding super (virtual) nodes in each layer and allowing input features to exchange information with them to absorb useful knowledge. The overview of our proposed module is illustrated in fig.~\ref{fig4}.

Specifically, given the input features $x\in\mathcal{R}^{C_{in}\times T\times N}$ in each layer, they first pass through a LayerNorm layer and a linear layer to enhance their representations. These features are then multiplied with the super node features $u\in \mathcal{R}^{C_{in}\times E}$ to obtain the similarity matrix $\tilde{A}\in \mathcal{R}^{E\times T\times N}$. Here, $E$ denotes the number of super nodes. To encode the importance of each super node, we introduce a randomly initialized learnable parameter $\beta \in \mathcal{R}^{E \times 1\times 1}$, and multiply it with the similarity matrix. The similarity matrix $\tilde{A}$ then passes through a tanh activation function to transform its values into [-1, 1], whose negative values would suppress corresponding connections and positive values would enhance the received messages. Finally, the super node features $u\in \mathcal{R}^{C_{in}\times E}$ are multiplied with the computed similarity matrix $\tilde{A}\in \mathcal{R}^{E\times T\times N}$ to aggregate beneficial information from super nodes for each body joint, to absorb specific domain knowledge to help recognition.

\subsection{Dynamic Graph Aggregation Module}
In addition to the dynamic adjacent matrix that is computed independently for each input, we also integrate static graphs into our model design, which are invariant over inputs to encode inherently important connections between body points. Practically, we introduce a fully learnable graph $P\in \mathcal{R}^{C_{in}\times N\times N}$ to capture joint relationships. $P$ is randomly initialized and updated via backward-gradient-based network propagation, which is channel-wisely set to learn unique connection characteristics for each channel to encode body movements. To exchange messages between different graph nodes, we multiple input features $x\in\mathcal{R}^{C_{in}\times T\times N}$ with $P\in \mathcal{R}^{C_{in}\times N\times N}$ to let each node aggregate information from all other joints in a per-channel manner, resulting in output features $y \in \mathcal{R}^{T\times C_{in}\times N}$.

\subsection{Parallel Temporal Convolution Module}
Current techniques typically utilize a single temporal convolutional module with a large kernel to extract temporal information. Nonetheless, we argue that this approach may not be adequate to represent the complex movements when performing signs. For instance, various signers may perform signs with distinct speeds and body movements, and the action durations may vary significantly. Using a single temporal receptive field to capture such diverse movement patterns may not be optimal. Therefore, we propose a parallel temporal convolution module (PTCN) that consists of multiple temporal convolutions in parallel with different kernel sizes to capture these dynamics. 

\begin{figure}[t]
  \centering
  \includegraphics[width=0.8\columnwidth]{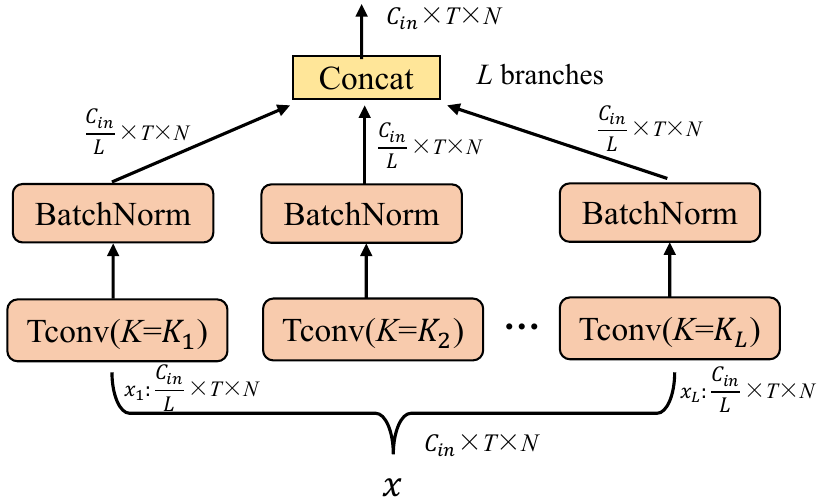} 
  \caption{Overview for the proposed parallel temporal convolution module.} 
  \label{fig5}
\end{figure}

Fig.~\ref{fig5} depicts the architecture of our proposed module. It comprises $L$ parallel branches with various kernel sizes, i.e., {$K_1$, $K_2$, $\dots$, $K_L$}, to capture temporal information of various scales. Given the input features $x\in \mathcal{R}^{C_{in}\times T\times N}$, we split $x$ along the channel dimension into $L$ branches of {$x_1$, $\cdots$, $x_L$}$\in\mathcal{R}^{C_{in}/L\times T\times N}$ independently. In each of the $L$ branches, such as the $l_{th}$ branch, $x_l$ undergoes a temporal convolution with a kernel size of $K_l$ to capture temporal information. Then, a BatchNorm layer calibrates the representations. Finally, the features from different branches are concatenated along the channel dimension as outputs to mix information from different scales. 

As shown in fig~\ref{fig2}, we serially stack three PTCNs to increase the temporal receptive field and enlarge the model capacity to capture complex human dynamics.

\subsection{Multiple-Stream Fusion.}
Following previous methods~\cite{jiang2021sign,jiang2021skeleton}, we fuse beneficial information from multiple streams to perform recognition, including joint (original input data), bone, joint-motion and bone motion streams. For the bone stream, we obtain information by subtracting the source joint (which is close to the center of gravity of the skeleton) from the target joint (which is far away from the center of gravity) for each connected joint pair of the human body. Similarly, we obtain motion information by subtracting the difference between adjacent frames for either joint or bone streams. We calculate softmax scores for all four streams and then average them to give the final prediction.

\section{Experiments}
\subsection{Experimental Setup}
\subsubsection{Datasets}

\textbf{SLR500~\cite{zhang2016chinese}} is a Chinese sign language dataset recorded in the controlled lab environment with a solid-color background. It contains 500 words performed by 50 signers through 5 times. Totally, there are 125,000 videos in SLR500. 

\textbf{WLASL~\cite{li2020word}} is a challenging American Sign Language dataset collected from web videos with a vocabulary size of 2000 words. It contains 21,083 samples performed by 119 signers with unconstrained recording conditions. We follow previous methods to use four subsets of WLASL, i.e., WLASL100, WLASL300, WLASL1000 and WLASL2000 to evaluate our method.

\textbf{MSASL~\cite{joze2018ms}} is an American sign language dataset, consisting of 16,054, 5,287, and 4,172 samples in the training, testing, and validation set, respectively. It's recorded in unconstrained real-life scenarios with a vocabulary size of 1,000. We follow previous methods to use four subsets of MSASL, i.e., MSASL100, MSASL200, MSASL500 and MSASL1000 to evaluate our method.

\textbf{NMFs-CSL~\cite{hu2021global}} contains 25,608 and 6,402 samples for training and testing, respectively, with a vocabulary of 1067 words. 

\subsubsection{Training details}
We use $r$=4 basic blocks, and set the channels for each block in [64, 128, 256, 512]. We randomly/centrally sample 120 frames out of 150 input frames as input during training/testing. Input data is mean normalized, added with random noise of maxmium value of 20 and augmented with 50\% horizontally flipping (only during training). The temporal stride of 3rd PTCN in each block is set as 2 to decrease the sequence length. We adopt the SGD optimizer to train our model by total 250 epochs and learning rate 0.1, which is decreased by a factor of 10 after 150 and 200 epochs. Our model is trained with batch size of 24 with weight decay 0.0001 on a 3090 GPU in PyTorch framework. We follow SAM-SLR-v2~\cite{jiang2021sign} to add a STC Module after each block to calibrate the input features. 

\subsection{Ablation Study}
We conduct experiments on the WLASL2000 dataset with joint stream only to verify the effectiveness of our proposed modules.

\begin{table}
  \centering
  \begin{tabular}[t]{lc}
  \toprule
  \textbf{\makecell[l]{Model configurations}}  & \textbf{Top-1(\%)} \\
  \midrule
  Our model & \textbf{51.44} \\ 
  \quad w/o Graph correlation &  49.25 (-2.19) \\ 
  \quad w/o Super node transform &  50.82 (-0.62) \\ 
  \quad w/o Dynamic aggregation &  50.23 (-1.21) \\ 
  \quad w/o PTCN &  49.69 (-1.75) \\ 
  \bottomrule    
  \end{tabular}
  \caption{Ablations on the effectiveness of each proposed module.}
  \label{tab1}
  \end{table}

\begin{table}
  \centering
  \begin{tabular}[t]{ccc}
  \toprule
  \textbf{\makecell[c]{Model configurations}}  & \textbf{Top-1(\%)} \\
  \midrule
  GCN &  47.51\\
  Learnable graph & 49.86\\
  Graph correlation module  & \textbf{51.44} \\ 
  \midrule
  $S$ = 1 & 50.64\\
  $S$ = 4 & 50.91\\
  $S$ = 8 & \textbf{51.44} \\
  $S$ = 16 & 50.93 \\
  \bottomrule    
  \end{tabular}
  \caption{Ablations on the configuration of graph correlation module.}
  \label{tab2}
  \end{table}  
\textbf{Study on the effectiveness of proposed modules.} Tab.~\ref{tab1} verifies the effectiveness of our proposed modules with an overall 51.44\% accuracy on the WLASL2000 dataset. It has been observed that eliminating any of the suggested modules leads to a decrease in accuracy. Especially, it's noticed that the graph correlation module and PTCN have the greatest impact on accuracy, resulting in a boost of +2.19\% and +1.75\%, respectively, by distinguishing beneficial spatial features from other nodes or capturing complex human dynamics.

\textbf{Study on the architecture of graph correlation module.} In the upper part of tab.~\ref{tab2}, we compare our proposed graph correlation module against other spatial aggregation architectures, e.g., GCN and a learnable graph. GCN aggregates spatial information from 1-hop neighbors with fixed weights. The learnable graph uses an adaptive adjacent matrix for all inputs. It's noticed that our proposed graph correlation module outperforms the other two architectures by a large margin in accuracy. We then examine the effects of the number of generated adjacent matrices $S$, and find that larger values of $S$ lead to better performance, reaching a peak at 8. We thus set $S$=8 by default. 

  \begin{table}[t]
    \centering
    \begin{tabular}[t]{ccc}
    \toprule
    \textbf{\makecell[c]{Model configurations}}  & \textbf{Top-1(\%)} \\
    \midrule
    $E$ = 1 & 50.98\\
    $E$ = 6 & \textbf{51.44}\\
    $E$ = 12 & 51.26\\
    $E$ = 24 & 50.84\\
    \midrule
    w/o $\beta$ & 50.96\\
    Tanh $\rightarrow$ softmax & 50.34\\
    \bottomrule    
    \end{tabular}  
    \caption{Ablations on the configuration of super node transform module.}
    \label{tab3}
    \end{table} 

\begin{table}[t]
    \centering
    \begin{tabular}[t]{ccc}
    \toprule
    \textbf{\makecell[c]{Model configurations}}  & \textbf{Top-1(\%)} \\
    \midrule
    Kernels = [3,5,7,9] & 50.05\\
    Kernels = [3,5] & 50.22\\
    Kernels = [3,9] & 51.05\\
    Kernels = [5,7] & \textbf{51.44}\\
    Kernels = [7,9] & 51.03\\
    \midrule
    1*PTCN & 49.64\\
    2*PTCN & 50.52\\
    3*PTCN & \textbf{51.44}\\
    4*PTCN & 50.66\\
    \bottomrule    
    \end{tabular}  
    \caption{Ablations on the configuration of parallel temporal convolution module.}
    \label{tab4}
    \end{table}   

\begin{table*}
  \centering
  \setlength\tabcolsep{3pt}
  \begin{tabular}[t]{ccccccccc}
  \toprule
  \multirow{2}{*}{Methods} & \multicolumn{2}{c}{WLASL100} & \multicolumn{2}{c}{WLASL300} & \multicolumn{2}{c}{WLASL1000} & \multicolumn{2}{c}{WLASL2000} \\
  & P-I(\%) & P-C(\%) & P-I(\%) & P-C(\%) & P-I(\%) & P-C(\%) & P-I(\%) & P-C(\%) \\
  \midrule
  \textbf{Skeleton-aware} \\
  SignBERT~\cite{hu2021signbert} & 79.07 & 80.05 & 70.36 & 71.17 & - & - & 47.46 & 45.17\\
  HMA~\cite{hu2021hand} & - & - & - & - & - & - & 46.32 & 44.09\\
  SAM-CLR-v2~\cite{jiang2021sign}& - & - & - & - & - & - & 51.50 & -\\
  SLGTformer~\cite{song2022slgtformer} & - & - & - & - & - & - & 47.42 & -\\
  BEST~\cite{zhao2023best} & 77.91 & 77.83 & 67.66 & 68.31 & - & - & 46.25 & 43.52\\
  \midrule
  \textbf{RGB-based} \\
  I3D~\cite{li2020word} & 65.89 & - & 56.14 & - & 47.33 & - & 32.48 & - \\
  Fusion-3~\cite{hosain2021hand} & 75.67 & - & 68.30 & - & 56.68 & -&  38.84 & - \\
  TCK~\cite{li2020transferring} & 77.52 & - & 68.56 & - & - & - & - & -\\
  BEST~\cite{zhao2023best} (R+P) & 81.63 & 81.01 & 75.60 &  76.12 & - & - & \textbf{54.59} & \textbf{52.12} \\
  \midrule  
  Ours & \textbf{82.38} & \textbf{83.09} & \textbf{79.97} & \textbf{80.56} & \textbf{67.76} & \textbf{67.54} & 53.68 & 51.17\\
  \bottomrule
  \end{tabular}  
  \caption{Comparison with other methods on the WLASL dataset. 'R+P' denotes the fused results of RGB and pose modalities.}
  \label{tab5}
  \end{table*}

  \begin{table*}
    \centering
    \setlength\tabcolsep{3pt}
    \begin{tabular}[t]{ccccccccc}
    \toprule
    \multirow{2}{*}{Methods} & \multicolumn{2}{c}{MSASL100} & \multicolumn{2}{c}{MSASL200} & \multicolumn{2}{c}{MSASL500} & \multicolumn{2}{c}{MSASL1000} \\
    & P-I(\%) & P-C(\%) & P-I(\%) & P-C(\%) & P-I(\%) & P-C(\%) & P-I(\%) & P-C(\%) \\
    \midrule
    \textbf{Skeleton-aware} \\
    SignBERT~\cite{hu2021signbert} & 81.37 & 82.31 & 77.34 & 78.02 & - & - & 59.80 & 57.06\\
    HMA~\cite{hu2021hand} & 78.57 & 79.48 & 72.19 & 73.52 & - & - & 56.02 & 52.98\\
    BEST~\cite{zhao2023best} & 80.98 & 81.24 & 76.60 & 76.75 & - & - & 58.82 & 54.87\\
    \midrule
    \textbf{RGB-based} \\
    I3D+BLSTM~\cite{li2020word} & 72.07 & -& -& -& -& -& 40.99 & - \\
    TCK~\cite{li2020transferring} & 83.04 & 83.91 & 80.31 & 81.14 & - & - & - & -\\
    BSL~\cite{albanie2020bsl} & - & - & - & - & - & - & 64.71 & 61.55\\
    \midrule  
    Ours & \textbf{84.16} & \textbf{84.54} & \textbf{81.58} & \textbf{81.89} & \textbf{75.06} & \textbf{74.68} & \textbf{65.74} & \textbf{62.31}\\
    \bottomrule
    \end{tabular} 
    \caption{Comparison with other methods on the MSASL dataset. 'R+P' denotes the fused results of RGB and pose modalities.}
    \label{tab6} 
    \end{table*}

\begin{table}
    \centering
    \begin{tabular}[t]{ccc}
    \toprule
    Methods  & \textbf{Top-1(\%)} \\
    \midrule
    \textbf{Skeleton-aware} \\
    ST-GCN~\cite{yan2018spatial} &  90.0\\
    SignBERT~\cite{hu2021signbert} & 96.6 \\
    HMA~\cite{hu2021hand} & 95.9 \\
    \midrule
    \textbf{RGB-based} \\
    3D-R50~\cite{qiu2017learning} &  95.1\\
    GLE-Net~\cite{hu2021global} & 96.8\\
    SignBERT~\cite{hu2021signbert} (R+P) & 97.6 \\
    BEST~\cite{zhao2023best} (R+P) & 97.7 \\
    \midrule
    Ours & \textbf{98.1}\\
    \bottomrule
    \end{tabular}
    \caption{Comparison with other methods on the SLR500 dataset. 'R+P' denotes the fused results of RGB and pose modalities.}
    \label{tab7}  
    \end{table} 

\begin{table}
  \centering
  \setlength\tabcolsep{3pt}
  \begin{tabular}[t]{ccc}
  \toprule
  Methods  & \textbf{Top-1(\%)} \\
  \midrule
  \textbf{Skeleton-aware} \\
  ST-GCN~\cite{yan2018spatial} & 59.9 \\
  HMA~\cite{hu2021hand} & 71.7\\
  SignBERT~\cite{hu2021signbert} & 74.9\\
  BEST~\cite{zhao2023best}  & 68.5\\
  \midrule
  \textbf{RGB-based} & \\
  I3D~\cite{carreira2017quo} & 64.4\\
  TSM~\cite{lin2019tsm} & 64.5\\
  Slowfast~\cite{feichtenhofer2019slowfast} & 66.3 \\
  GLE-Net~\cite{hu2021global} & 69.0\\
  HMA~\cite{hu2021hand} (R+P)& 75.6\\
  \midrule
  Ours & \textbf{77.9} \\
  \bottomrule
  \end{tabular}  
  \caption{Comparison with state-of-the-art methods on the NMFs-CSL dataset. 'R+P' denotes the fused results of RGB and pose modalities.}
  \label{tab8}
  \end{table}

\begin{figure*}[t]
  \centering
  \includegraphics[width=1.6\columnwidth]{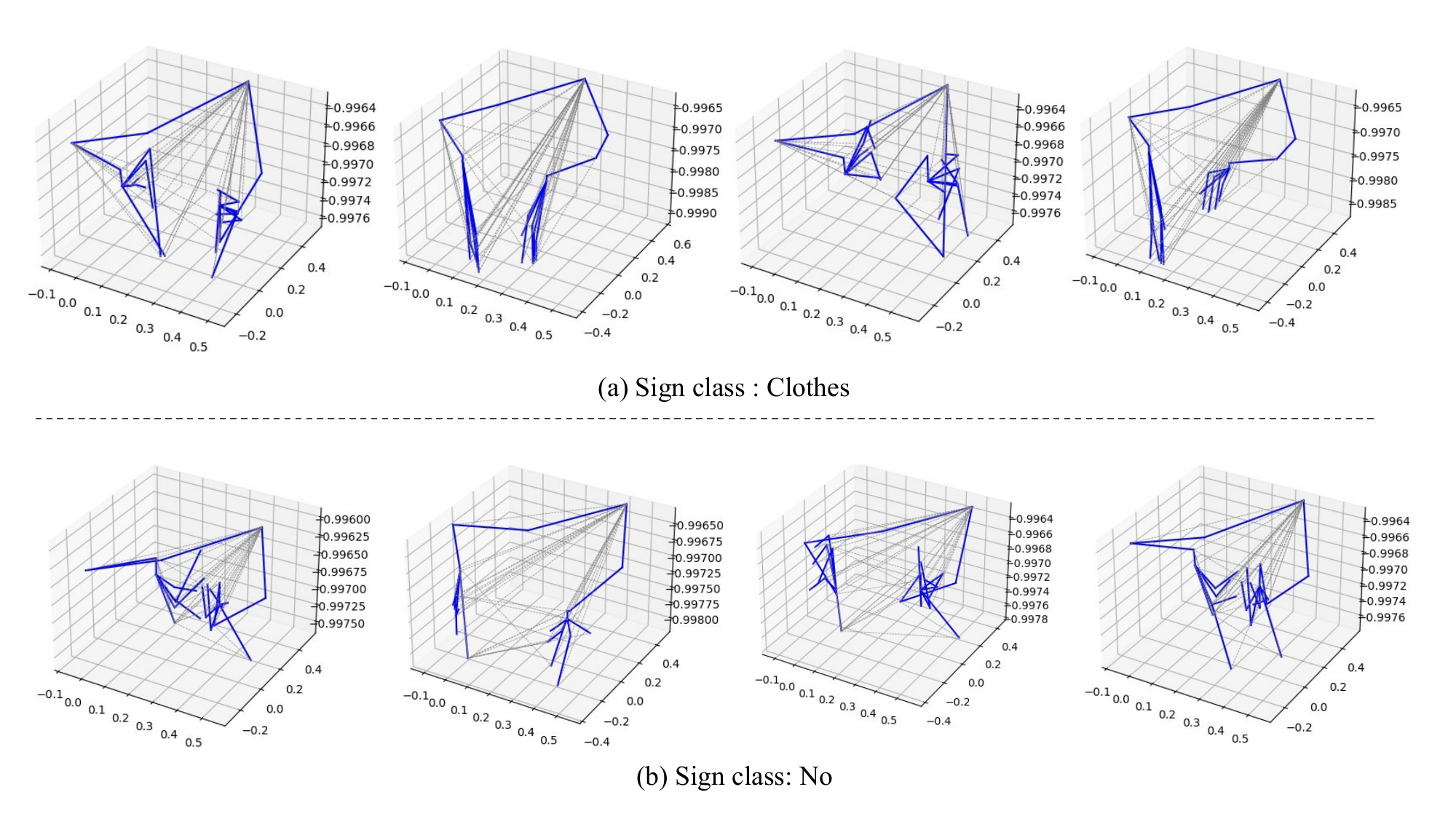} 
  \caption{Visualizations of learned edges in the graph correlation module for two signs. Only the top 5\% active connections are plotted. It's observed that our method learns to build dynamic joint relationships for different input samples, and especially pays attention to the joints of both hands to capture sign movements.} 
  \label{fig6}
\end{figure*}

\textbf{Study on the configurations of super node transform module.}  We first demonstrate the impact of varying the number of super nodes $E$. It's observed that larger $E$ achieves better performance, which reaches a peak after equalling 6. Therefore, we set $E$=6 by default. Next, we evaluate the effectiveness of several proposed components in the super node transform module. Removing $\beta$ leads to a decrease in accuracy, as it dynamically adjusts the importance weights of different super nodes. Additionally, replacing the tanh activation function with softmax results in decreased accuracy. The tanh activation function is crucial for highlighting important features with positive outputs and suppressing irrelevant features with negative outputs.  

\textbf{Study on the configurations of parallel temporal convolution module (PTCN).} The upper part of tab.~\ref{tab4} investigates the architecture of PTCN. We find that a two-branch architecture using kernel sizes of [5,7] achieves better performance than other options, such as two branches or four branches. We believe that larger temporal kernels (e.g., 9) may introduce excessive noise, while smaller kernels may not adequately capture complex temporal movements. We then consider the number of stacked PTCNs. The bottom part of tab.~\ref{tab4} reveals that increasing the number of PTCNs leads to consistent improvements in performance, which reaches a peak after using 3 PTCNs. Thus, we use 3 PTCNs as the default setting. 

\subsection{Comparison with the state-of-the-art}
\textbf{WLASL.} Tab.~\ref{tab5} presents a comparison of our method with state-of-the-art approaches on the WLASL dataset. For fair comparison, we follow the previous methods to divide this dataset into four subsets, including WLASL100, WLASL300, WLASL1000, and WLASL2000, and evaluate using two accuracy metrics: per-instance accuracy (P-I) and per-class accuracy (P-C). The upper part of Tab.~\ref{tab5} compares our method with other skeleton-aware methods, while the bottom part compares it with other RGB-based methods. Our method achieves new state-of-the-art performance across all four subsets, surpassing other skeleton-aware approaches by a large margin, thanks to its superior ability to dynamically aggregate joint features and capture complex human dynamics. We especially find that our method outperforms other RGB-based approaches in most cases, despite consuming fewer computation. These results demonstrate the effectiveness and efficiency of our proposed method.

\textbf{MSASL.} Tab.~\ref{tab6} compares our method with state-of-the-art approaches on the MSASL dataset, which is divided into four subsets including MSASL100, MSASL200, MSASL500 and MSASL1000. The upper and bottom parts of tab.~\ref{tab6} compare our method against other skeleton-aware methods and RGB-based methods, respectively. It's observed that our method achieves much better accuracy in both metrics across all four subsets than other skeleton-aware approaches. Moreover, our method outperforms RGB-based approaches in most cases.

\textbf{SLR500 and NMFs-CSL.} Tab.~\ref{tab7} and tab.~\ref{tab8} compare our method against other skeleton-aware and RGB-based methods on the SLR500 and NMFs-CS datasets, respectively. As a skeleton-aware approach, our method beats both skeleton-aware methods and RGB-based methods, verifying its effectiveness.

\subsection{Visualizations}
Fig.~\ref{fig6} depicts the edges computed by the graph correlation module to demonstrate its efficacy in capturing human dynamics for the "Clothes" and "No" signs. Only the edges with top 5\% weights are displayed. It is observed that the generated edges primarily connect the center joint (neck) with joints located in both hands for both signs. Additionally, our approach learns to establish connections between the joints of the left and right hands to create distant relationships. We conclude that our method can develop dynamic joint relationships for various input samples and, in particular, focuses on the joints of both hands to capture sign movements.

\subsection{Efficiency}
\subsubsection{Model parameters}
Tab.~\ref{tab9} presents a comparison between our proposed method and two recent skeleton-aware approaches, namely SAM-CLR-v2~\cite{jiang2021sign} and SPOTER~\cite{bohavcek2022sign}. We observe that while the parameters of our method fall between those of the other approaches, our method significantly outperforms them in terms of accuracy. This demonstrates the effectiveness and efficiency of our proposed approach.

\subsubsection{Comparison with RGB-based methods}
\begin{figure}[t]
  \centering
  \includegraphics[width=\columnwidth]{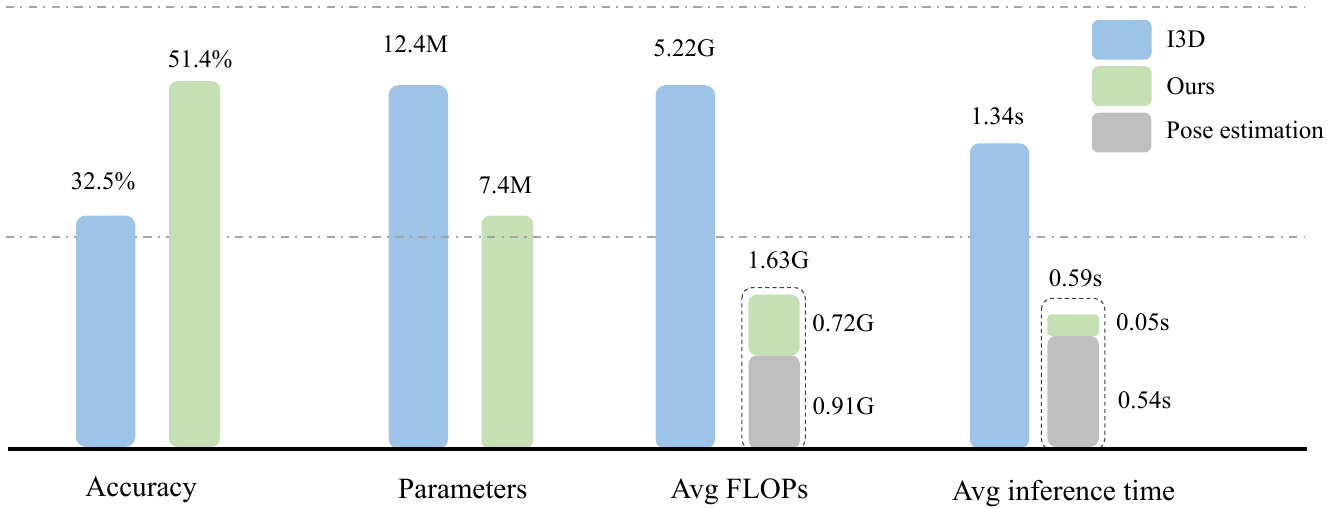} 
  \caption{Comparison of our method (with joint stream only) and I3D~\cite{li2020word} upon accuracy, parameters, average FLOPs, and average inference time on the WLASL2000 dataset.}
  \label{fig7}
\end{figure}

Fig.~\ref{fig7} presents a comparison between our method and a popular RGB-based approach, I3D~\cite{li2020word}, using various metrics such as accuracy, parameters, FLOPs, and inference time on the WLASL2000 dataset to highlight the advantages of our method in real-world applications. The figure shows that our method outperforms I3D in terms of accuracy, while having lower parameters, much fewer average FLOPs per video, and less average inference time per video. Compared to RGB-based approaches, our method demonstrates great advantages in both accuracy and efficiency by taking skeleton data as input, demonstrating its superiority in real-life scenarios.

\begin{table}
  \centering
  \begin{tabular}[t]{ccc}
  \hline
  Methods  & Parameters(M) & Top-1(\%)\\
  \hline
  SAM-CLR-v2 & 4.3 & 45.61\\
  SLGTformer & 9.3 & 47.42 \\
  \hline
  Ours & 7.4 & \textbf{51.44}\\
  \hline
  \end{tabular}  
  \caption{Comparison of our method with other skeleton-aware approaches upon both accuracy and parameters on the WLASL2000 dataset with joint stream only.}
  \label{tab9}
  \end{table} 

\section{Conclusion}
In this paper, we address the shortcomings of previous skeleton-aware sign language recognition methods by introducing two novel approaches. Firstly, we dynamically construct joint relationships to gather useful spatial features. Secondly, we present a parallel temporal convolution module to capture intricate human dynamics. Our method achieves new state-of-the-art results on four widely-used benchmarks and even outperforms RGB-based approaches in terms of both effectiveness and efficiency in most cases.
\section{Acknowledgements}
This study was supported by National Key Research and Development Program of China (2023YFF0906200) and National Natural Science Foundation of China (Grant Nos. 62072334).
\bibliographystyle{lrec-coling2024-natbib}
\bibliography{ref}

\end{document}